\documentclass[letterpaper, 10 pt, conference]{ieeeconf}  

\IEEEoverridecommandlockouts                              

\usepackage{amsmath} 
\usepackage{amssymb}  
\usepackage{multirow}
\usepackage{graphicx}
\usepackage{booktabs}
\usepackage{tabularx}
\usepackage{hyperref}
\usepackage{bbm}
\usepackage{dblfloatfix}
\usepackage{authblk}
\usepackage{cite}

\hypersetup{
    colorlinks=true,
    urlcolor=blue,
}
\DeclareMathOperator{\E}{\mathbb{E}}
\DeclareMathOperator*{\argmax}{arg\,max}

\newcommand{\mathds}[1]{\text{\usefont{U}{dsrom}{m}{n}#1}}

\title{\LARGE \bf

Tidiness Score-Guided Monte Carlo Tree Search for 

Visual Tabletop Rearrangement 

}

\author{Hogun Kee, Wooseok Oh, Minjae Kang, Hyemin Ahn, and Songhwai Oh
\thanks{H. Kee, W. Oh, M. Kang and S. Oh are with the Department of Electrical and Computer Engineering and ASRI, 
		Seoul National University, Seoul 08826, Korea
		(e-mail: {\tt\small \{hogun.kee, wooseok.oh, minjae.kang\}@rllab.snu.ac.kr, songhwai@snu.ac.kr}).
        }
\thanks{
        H. Ahn is with Artificial Intelligence Graduate School (AIGS), Ulsan National Institute of Science and Technology (UNIST), Ulsan, Korea
        (e-mail:{ \tt\small hyemin.ahn@unist.ac.kr}).}
}

\begin{document}

\maketitle
\thispagestyle{empty}
\pagestyle{empty}

\begin{abstract}
In this paper, we present the tidiness score-guided Monte Carlo tree search (TSMCTS), a novel framework designed to address the tabletop tidying up problem using only an RGB-D camera.
We address two major problems for tabletop tidying up problem: 
(1) the lack of public datasets and benchmarks, and 
(2) the difficulty of specifying the goal configuration of unseen objects.
We address the former by presenting the tabletop tidying up (TTU) dataset, a structured dataset collected in simulation. 
Using this dataset, we train a vision-based discriminator capable of predicting the tidiness score.
This discriminator can consistently evaluate the degree of tidiness across unseen configurations, including real-world scenes.
Addressing the second problem, we employ Monte Carlo tree search (MCTS) to find tidying trajectories without specifying explicit goals. 
Instead of providing specific goals, we demonstrate that our MCTS-based planner can find diverse tidied configurations using the tidiness score as a guidance. 
Consequently, we propose TSMCTS, which integrates a tidiness discriminator with an MCTS-based tidying planner to find optimal tidied arrangements.
TSMCTS has successfully demonstrated its capability across various environments, including coffee tables, dining tables, office desks, and bathrooms. 
The TTU dataset is available at: \url{https://github.com/rllab-snu/TTU-Dataset}.

\end{abstract}

\section{Introduction}
\begin{figure}[htb!]
\begin{center}
\includegraphics[width=0.48\textwidth]{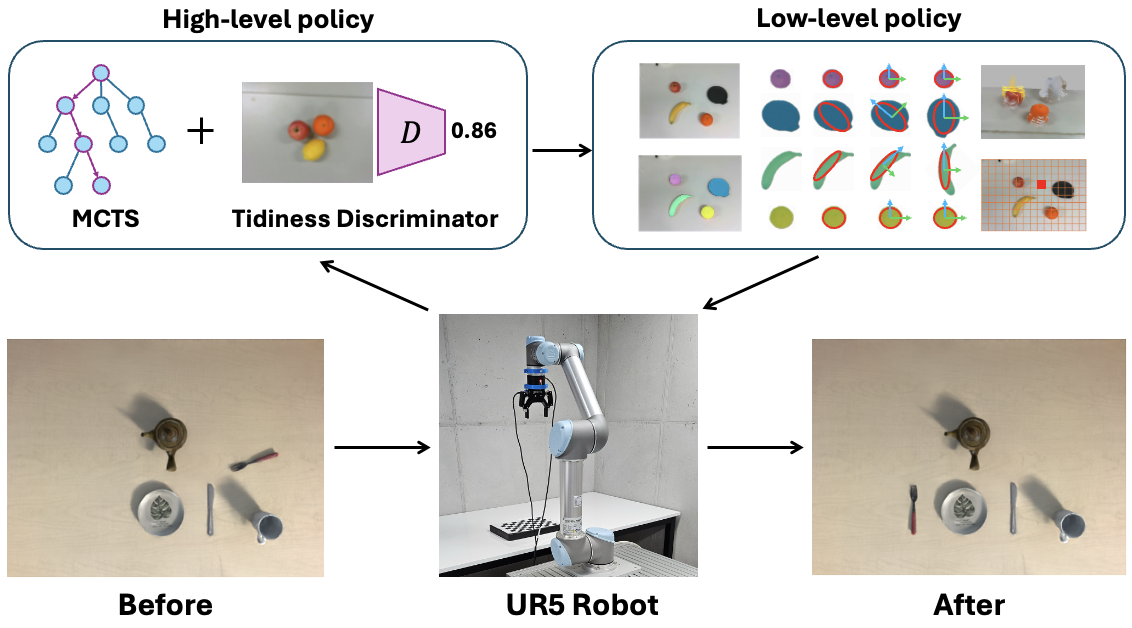}
\end{center}
\vspace*{-4mm}
\caption{
The hierarchical policy of TSMCTS iteratively finds pick-and-place actions to tidy up objects on a table.
The high-level policy finds which object to pick and place according to the current configuration. 
The low-level policy finds grasp points and trajectories of the end effector.
Details of each policy are described in Section V.
}
\vspace*{-4mm}
\label{fig:tidying_problem}
\end{figure}

In this paper, we address the tabletop tidying problem, where an embodied AI agent autonomously organizes objects on a table based on their composition. 
As depicted in Figure \ref{fig:tidying_problem},
tidying up involves rearranging objects by determining an appropriate configuration of given objects, without providing an explicit target configuration.
Previous research has encountered difficulties in defining the tidying up problem, primarily due 
to the lack of public datasets and metrics to assess tidiness.
To address these issues, we collect a structured dataset for tabletop tidying, and train a tidiness discriminator and tidying planner to transform a messy table into an organized one through a simple and effective framework. 
We refer to this as the tidiness score-guided Monte Carlo tree search (TSMCTS).

In previous research on object rearrangement, goal configurations are provided either as target positions or as images of the desired arrangement \cite{large-scale, nerp, ifor}. 
This setup enables straightforward evaluation by comparing the state to the goal.
However, using an image as a goal requires objects to be pre-arranged for goal generation, limiting flexibility. 
To accommodate a wider variety of goals, recent studies employ language labels or descriptions
to define more abstract goals
\cite{StructFormer, StructDiffusion, LGMCTS}.
Nonetheless, these language-conditioned rearrangement studies require 
separate encoding modules to integrate language with image or positional inputs. 
To connect these domains, models like CLIP \cite{clip} and BLIP \cite{blip} map features into a unified latent space. 
To connect the language domain to the image domain, CLIP \cite{clip} and BLIP \cite{blip} models focus on mapping features from different domains into a unified latent space.
While these approaches show promising performance in extracting semantic features \cite{cliport, xiao2022robotic}, they still struggle with understanding the spatial relationships among objects. 

The proposed method, TSMCTS, learns a tidiness score function and finds an action sequence that generates a tidy configuration based on object combinations without explicit goals.
We collect a tabletop tidying dataset from diverse environments (e.g., coffee tables, dining tables, office desks, bathrooms) and train a tidiness discriminator to measure the degree of tidiness reliably, even with unseen objects and real-world images. 
Our discriminator is superior to previous approaches since it can consistently measure tidiness from real-world images, whereas previous attempts \cite{SceneScore, TarGF} primarily demonstrate success in simulations or toy examples rather than real-world scenarios.
Finally, we use the tidying-up discriminator as a utility function of MCTS \cite{mcts} to find a sequence of pick-and-place actions.
The proposed MCTS-based planner employs a tidying policy trained with our tabletop tidying-up dataset using an offline reinforcement learning method, specifically Implicit Q-learning (IQL) \cite{iql}, as its tree policy.

The proposed method, TSMCTS, achieves a tidying success rate of 88.5\% in simulation experiments and 85\% in real robot experiments. These results experimentally demonstrate TSMCTS’s ability to tidy up various combinations of objects using both simulations and real robots. 
We also perform a human evaluation, which demonstrates that the proposed method can tidy up a table as much as humans can perceive it well-arranged.

\section{RELATED WORK}
Tidying up is an object rearrangement problem occurring in situations where the goal is not explicitly provided. 
Instead of a specific goal, several research approaches involve expressing goals in natural language \cite{StructFormer, StructDiffusion, LGMCTS}, 
or finding functional arrangements based on user preferences \cite{myhouse, RobotOrganize}.
Additionally, there are studies that directly learn the degree of tidiness as a score function and plan trajectories to achieve a tidied scene \cite{SceneScore, TarGF}.

Recent studies such as StructFormer \cite{StructFormer} and StructDiffusion \cite{StructDiffusion} find appropriate positions for objects guided by natural language instructions. 
Both methods take language tokens and object point clouds as inputs to find arrangements that satisfy language conditions. 
Studies such as \cite{myhouse} and \cite{RobotOrganize} learn user preferences to find organized arrangements without explicit goals.
For instance, \cite{myhouse} uses scene graphs to encode scenes and learns user preference vectors, 
and \cite{RobotOrganize} addresses tasks involving the organization of various items into containers or shelves, learning pairwise preferences of objects. 
These studies rely more on semantic information rather than visual information of the objects.
There exist diffusion based methods to directly generate final arrangement images \cite{DalleBot}, \cite{LVDiffuser}.
These studies rely on the commonsense knowledge inherent in large language models (LLMs) and vision language models (VLMs) to find arrangements that are similar to human intentions. 

Similar to the current work, studies such as \cite{SceneScore} and \cite{TarGF} learn to quantify the degree of tidiness with a score function. 
\cite{SceneScore} uses an energy-based model to learn and predict the cost, which is most relevant to our work. 
While \cite{SceneScore} focused on finding positions for just one missing object, our study plans to find the optimal state by moving all movable objects on a table. 
\cite{TarGF} learns a score function to calculate the likelihood with the target distribution for each task, using this score to learn a policy for rearranging objects. 
Each task requires a separate target distribution, and the score function is trained separately for each task, whereas our study uses a single score function to tidy up across various environments.

Language-guided Monte-Carlo tree search (LGMCTS) \cite{LGMCTS} uses the MCTS algorithm to find trajectories to obtain arrangements that satisfy language conditions. 
LGMCTS assumes that explicit spatial conditions can be derived based on language conditions. 
They first establish these spatial conditions and then find a trajectory that arranges the objects to satisfy all these conditions. 
In this paper, we propose an algorithm that learns a score function to find various tidied arrangements without the guidance of language.

\section{Problem Formulation}
In the tabletop tidying up problem, an agent (in our case, a robot) $\mathcal{M}$ is tasked to rearrange a set of movable objects $O=\{o_1, o_2, \ldots, o_N\}$ to achieve a tidy arrangement.

At each timestep, the agent receives a single top-down view RGB-D image from a fixed overhead camera. The workspace is planar and all objects are assumed to be rigid bodies.
The robot $\mathcal{M}$ can interact with objects through pick-and-place actions to perform arbitrary translation and rotation changes.

We assume there is a tidiness-score function $\Psi$ which returns the degree of tidiness given an image of the tabletop with objects. 
This function assesses whether objects on the table are visually tidied up,
considering the types, shapes, and sizes of the objects, and assigns a tidiness score between 0 and 1, where 0 represents a completely messy scene and 1 indicates a well-arranged scene.

For a given set of objects $O$, the visual observation depends on the pose of the objects.
Therefore, we can formulate the tidiness score as $\psi = \Psi(O, P)$, where $P$ denotes the 6-DoF positions of $O$.
Then we can formulate the objective of the table tidying problem as finding the optimal arrangement $P^*$ to maximize the tidiness score:
\begin{align}
P^* = \underset{P}{\argmax}\, \Psi(O, P) .
\end{align}

In this paper, we parameterize a tidiness-score function with neural networks $\theta$, as a discriminator $\Psi_\theta$.
$\Psi_\theta$ is trained to estimate the tidiness-score of a tabletop arrangement image.

\section{Tabletop Tidying Up Dataset}
We collect a Tabletop Tidying Up (TTU) dataset which includes both tidied and messy scenes to train a vision-based tidiness discriminator.
To cover diverse object arrangements,
we define a set of environments \(E\) consisting of four environments: Coffee table, Dining table, Office desk, and Bathroom. 
For each environment \(e \in E\), we define a set of objects \(O_e \subseteq O_\textrm{all}\) belonging to that environment, where \(O_\textrm{all}\) is the entire set of objects.
Then, we predefine possible combinations of objects within \(O_e\), each consisting of two to nine objects.

\begin{figure}[t!]
\begin{center}
\includegraphics[width=0.43\textwidth]{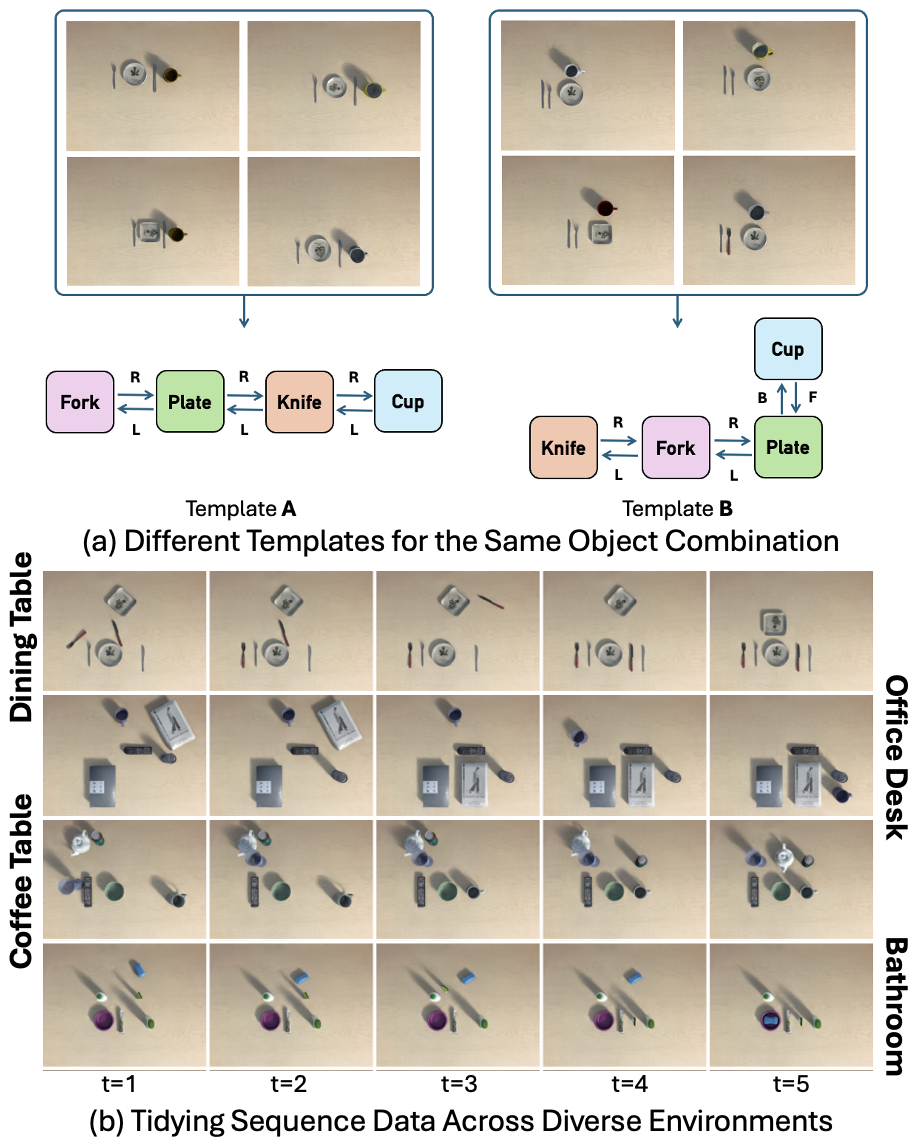}
\end{center}
\vspace*{-4mm}
\caption{
(a) Different arrangements can be created with the same combination of objects. 
R represents `right', L stands for `left', B for `behind', and F denotes `front' among the spatial relations.
Various templates are collected to capture as many tidied arrangements as possible for each object set.
(b) The TTU dataset consists of state-action sequences for each environment, ranging from a messy scene (t=1) to a perfectly tidied scene (t=T).
}
\vspace*{-1mm}
\label{fig:ttudataset}
\end{figure}

In this study, we introduce the concept of a template to encourage automatic collection of well-organized arrangement data.
A template is defined as a specific set of spatial relationships between objects, categorized as one of the following: {on, under, left, right, front, behind, left-front, left-behind, right-front, and right-behind}.
Figure \ref{fig:ttudataset}-a shows examples of templates and their corresponding tidied arrangements for a set of objects \(O=\{knife, fork, plate, cup\}\).
An arrangement where the fork is to the left of the plate and the knife to the right could all be considered as belonging to template $\mathcal{A}$.
Meanwhile, an arrangement where both the fork and knife are neatly placed on the left side of the plate would fall under template $\mathcal{B}$. 
By defining templates in this manner, we can represent all tidied arrangements that a person might create with specific templates. 

We first collect templates for the given object sets and then use these templates to gather tidied scene data.
We configure an appropriate combination of objects for each environment and design up to 16 templates for each combination of objects.
The entire process of finding templates is conducted manually by spawning object models on a table in the PyBullet simulation. 
We use 3D object models from the YCB dataset \cite{ycb} and the HouseCat6D dataset \cite{housecat}.

\begin{table}[t]
\centering
\caption{Data Collection Across Various Environments}
\label{tab:data_summary}
\vspace*{-1mm}
\begin{tabular}{@{}lcccc@{}}
\toprule
Environment & \# Objects & \# Templates & \# Trajectories & \# Data \\ \midrule
Coffee Table & 93 & 120 & 14,880 & 74,400 \\
Dining Table & 105 & 125 & 13,245 & 66,225 \\
Office Desk & 43 & 131 & 12,865 & 64,325 \\
Bathroom & 29 & 37 & 3,855 & 19,275 \\ \midrule
Total & 170 & 413 & 44,845 & 224,225 \\ \bottomrule
\end{tabular}
\vspace*{-6mm}
\label{table:data_collection}
\end{table}

To collect tidying sequence data, we first created tidied scenes based on templates, then generated untidying sequences by scattering the tidied objects one by one. 
After sampling a template, we create a tidied scene by augmenting the distances between objects, changing objects within the same category, and modifying the central position of the arrangements. 
This tidied scene becomes the final state $s_T$ of the trajectory, where $s_t$ represents the scene at timestep $t$, and $T$ denotes the trajectory length. 
We start from $s_T$ and randomly pick objects to move to random positions on the desk, collecting the untidying sequence of $(s_{T-1}, s_{T-2}, ..., s_1)$. 
By reversing this sequence, we obtain a tidying sequence from a messy to a tidied table. 
Finally, we collect a dataset $D=\{\tau_1, ..., \tau_{N_{traj}}\}$, where each trajectory $\tau_i = ((s_1, \psi_1), ..., (s_T, \psi_T))$ consists of a sequence of state and tidiness score pairs. 
The tidiness scores are given as follows, proportional to the timestep $t$, with the final state $s_T$ receiving a tidiness score of 1: 
\begin{align} 
\psi_t = \frac{t-1}{T-1} 
\end{align}
In Figure \ref{fig:ttudataset}-b, we collect tidying sequences using a trajectory length of $T=5$.
Table \ref{table:data_collection} lists the number of object models used in each environment, the number of templates, the number of trajectories, and the number of scene data.
There are overlapping objects across the environments, and we utilize a total of 170 object models.
In total, we have collected 413 templates and 224,225 scene data including RGB and depth images, object categories and 6-DoF positions of objects.

\begin{figure*}[htb!]
\begin{center}
\includegraphics[width=0.99\textwidth]{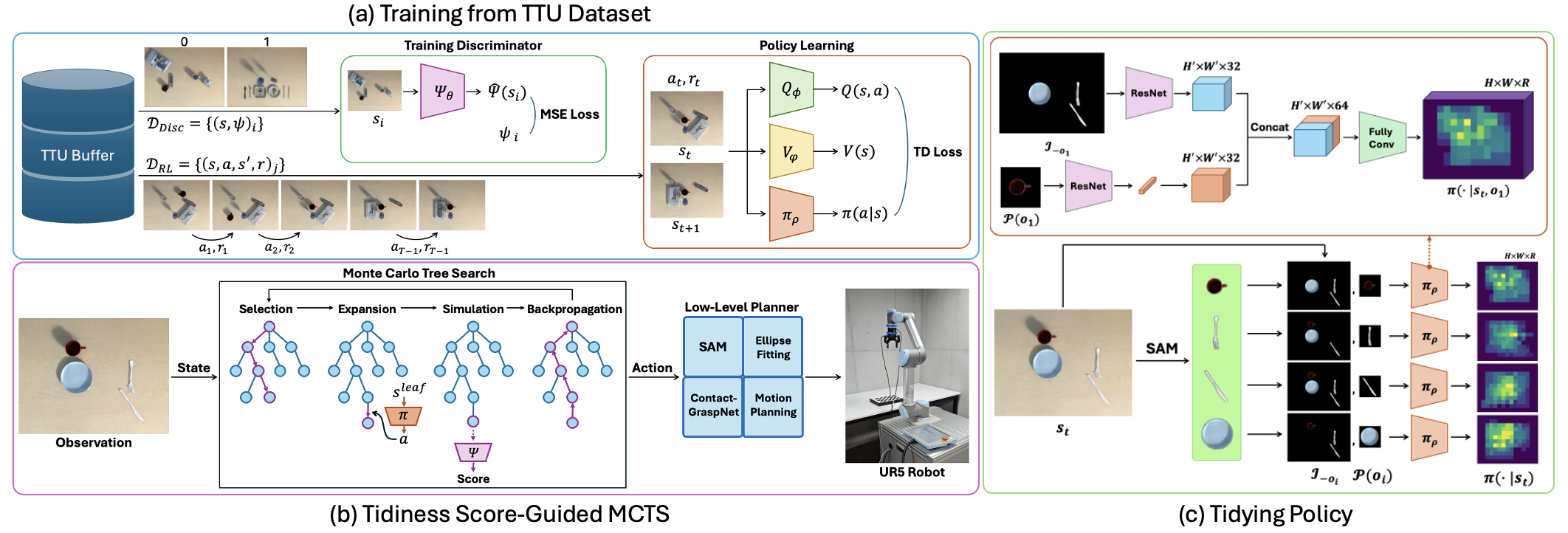}
\end{center}
\vspace*{-5mm}
\caption{
(a) We train the tidiness discriminator and tidying policy using the TTU dataset.
The tidiness discriminator is trained in a supervised manner to predict the tidiness score of the table, while
the tidying policy is trained to estimate the action distribution for pick-and-place actions using the IQL framework.
(b) During inference, MCTS utilizes the tidiness discriminator $\Psi_\theta$ and the tidying policy $\pi_\rho$ to find the best pick-and-place actions.
(c) From the current table image $s_t$, the policy networks take the table image \(\mathcal{I}_{-o_i}\) and the object's patch \(\mathcal{P}(o_i)\) as inputs to generate an action probability distribution. The action is defined by the selected object, its placement position, and its rotation.
}
\vspace*{-5mm}
\label{fig:overview}
\end{figure*}

\section{Proposed Method}
The proposed framework, the Tidiness Score-guided Monte Carlo Tree Search (TSMCTS), consists of two components: (1) training the tidiness discriminator and tidying policy, and (2) planning the tidying up process using MCTS.

We trained a tidiness discriminator and tidying policy using the TTU dataset described in the previous section. 
The tidiness discriminator learns a score function that evaluates the tidiness score of the current state. 
The tidying policy is used as a tree policy in the MCTS algorithm to efficiently sample appropriate actions from the entire feasible action space.
We trained the tidiness discriminator in a supervised manner and the tidying policy using the Implicit Q-Learning (IQL) framework.

Finally, starting from the initial configuration $(O, P)$, we iteratively find pick-and-place actions by planning with the MCTS algorithm using the tidiness discriminator as a utility function and the tidying policy as a tree policy, until all the objects on the table are tidied up.

\subsection{Tidiness Discriminator and Tidying Policy}
The training process of the tidiness discriminator and the tidying policy is illustrated in Figure \ref{fig:overview}-a.
We parameterized the tidiness score function using neural networks $\Psi_\theta: S\mapsto[0,1]$ and the tidying policy $\pi_\rho: S\mapsto A$, 
where $S$ and $A$ represent the state space and action space, respectively. 
In this paper, the state is represented as an RGB image of the table, while the action corresponds to a pick-and-place operation, defined by the target object, placement position, and rotation angle.

From the TTU dataset, we can obtain sequences of state and tidiness score pairs $((s_1, \psi_1), ..., (s_T, \psi_T))$, where $T$ denotes the length of collected trajectories. 
Here, $s_1$ is the most messy scene and $s_T$ is the final tidied up scene.
Finally, we train the discriminator using pairs of states and score labels, $\mathcal{D}_{Disc}=\{(s, \psi)_i\}$, employing the mean squared error as the loss function:
\begin{align}
L(\theta)=\E \big[ (\Psi_\theta(s_t)-\psi_t )^2 \big].
\end{align}
For policy training, we use the Implicit Q-Learning (IQL) method.
We use a sparse binary reward as below:
\begin{equation}
r_t=
\begin{cases}
  1, & \text{if the episode ends ($t=T$)}
  \\
  0, & \text{otherwise}
\end{cases}
\end{equation}
and obtain offline data $\mathcal{D}_{RL}=\{(s_t, a_t, s_{t+1}, r_t)\}$ from TTU dataset. In IQL method, we learns Q-function $Q_\phi$, value function $V_\varphi$ and the policy $\pi_\rho$ simultaneously.
The loss functions for $V_\varphi$ and $Q_\phi$ are computed according to the modified TD learning procedure in IQL,
\begin{align}
L_V(\varphi) = \E_{s,a}\big[L_2^{\tau} (Q_{\hat{\phi}}(s,a) - V_\varphi(s)) \big],
\end{align}
where $L_2^{\tau}(u)=|\tau-\mathds{1}(u<0)|u^2$ represents the expectile regression loss, with $\tau=0.7$ used as the default value.
$Q_{\hat{\phi}}$ is a target network, which is a lagged version of $Q_\phi$.
\begin{align}
L_Q(\phi) = \E_{s,a,s',r}\big[
(r + \gamma V_\varphi (s') - Q_\phi(s,a))^2 \big].
\end{align}
Then, the policy extraction step can be applied using advantage weighted regression:
\begin{small}
\begin{align}
L_\pi(\rho) = \E_{s,a}\big[
\exp(\beta (Q_{\hat{\phi}}(s,a)-V_\varphi(s))) \log \pi_\rho(a|s) \big],
\end{align}
\end{small}where $\beta$ denotes the inverse temperature.

We employ a pre-trained ResNet-18 as the backbone of the tidiness discriminator, replacing its final fully connected layer with one that predicts a single tidiness score.
The tidiness discriminator takes the current table image $s_t$ as input and outputs the corresponding tidiness score $\Psi_\theta(s_t)$.
We use the Segment Anything Model (SAM) \cite{sam} to remove the background to learn a more consistent score function.

For the tidying policy, two inputs are used for each object \(o_i\), \(i=1,\ldots,N\): 
(1) the patch image of the object, \(\mathcal{P}(o_i)\), 
and (2) the table image without the object, \(\mathcal{I}_{-o_i}\). 
The policy outputs a probability distribution over pixel positions and rotations for placing the target object. 
As shown in Figure \ref{fig:overview}-c, the policy networks extract separate features for the table \(\mathcal{F}(\mathcal{I}_{-o_i})\) and the object \(\mathcal{F}(\mathcal{P}(o_i))\) using ResNet-18 networks. 
The 32-dimensional object feature \(\mathcal{F}(\mathcal{P}(o_i))\) is then expanded to match the size of the table feature and concatenated with it to form the combined feature, \(cat(\mathcal{F}(\mathcal{I}_{-o_i}), \mathcal{F}(\mathcal{P}(o_i)))\).
This combined feature is processed through fully convolutional networks to generate an \(H \times W \times R\) probability distribution, where $R$ denotes the number of possible rotations. 
The tidying policy repeats this process for all \(N\) objects in parallel to produce an \(N \times H \times W \times R\) action probability distribution.

\subsection{Low-Level Planner}
\begin{figure}[tb!]
\begin{center}
\includegraphics[width=0.44\textwidth]{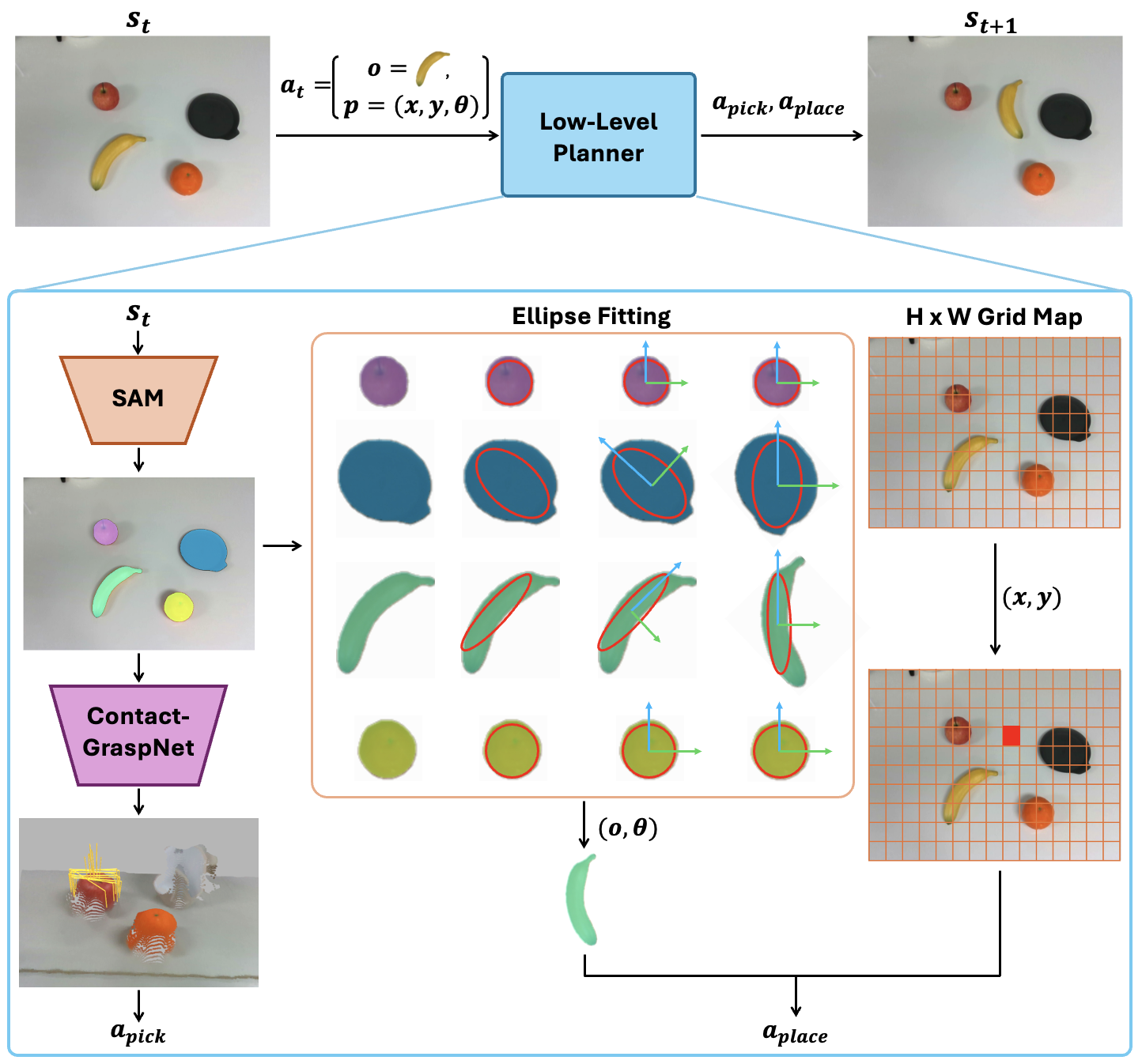}
\end{center}
\vspace*{-4mm}
\caption{
Given a high-level action specifying which object to pick and where to place, the low-level planner uses the Contact-GraspNet to find a stable grasping point for the object. To place the object in the desired orientation, the initial orientation is determined by applying ellipse fitting to the object mask obtained through SAM, followed by calculating the rotation transformation to determine the placement.
}
\vspace*{-6mm}
\label{fig:low-level}
\end{figure}

To discretize the pick-and-place action, we divide the workspace into an \(H \times W\) grid map and split the $360^\circ$~
rotation into \(R\) bins. Then, we define a pick-and-place action as \(a=(o, p)\), where \(o\) denotes the target object and \(p=(x,y,r)\) represents the placement position. Here, \(x\) and \(y\) are the selected pixel position, and \(r\) is the rotation index.
When picking up objects, we use Contact-GraspNet \cite{contact-graspnet}, which processes 3D point clouds from RGB-D images to output 6-DoF grasping points for the robot arm's end effector.
When placing objects, even if they are placed in the same location, they can appear tidy or completely disordered depending on the rotation of the objects. 
We observe that objects appear more organized to humans when aligned with the table’s $x$-axis or $y$-axis. 
To achieve this, we fit an ellipse to the object’s segmentation mask and use its major axis as the default rotation axis.
This process is illustrated in Figure \ref{fig:low-level}. 

First, the agent receives a top-down view RGB image of the table and uses SAM \cite{sam} to obtain a segmentation mask for each object.
Next, we use the least squares method to find an ellipse that fits each object and determine its major and minor axes.
The robot's placement action allows the object
to be aligned with the table's horizontal or vertical axis.

\subsection{Tidiness Score-Guided High-Level Planner} 
\begin{figure}[t!]
\begin{center}
\includegraphics[width=0.44\textwidth]{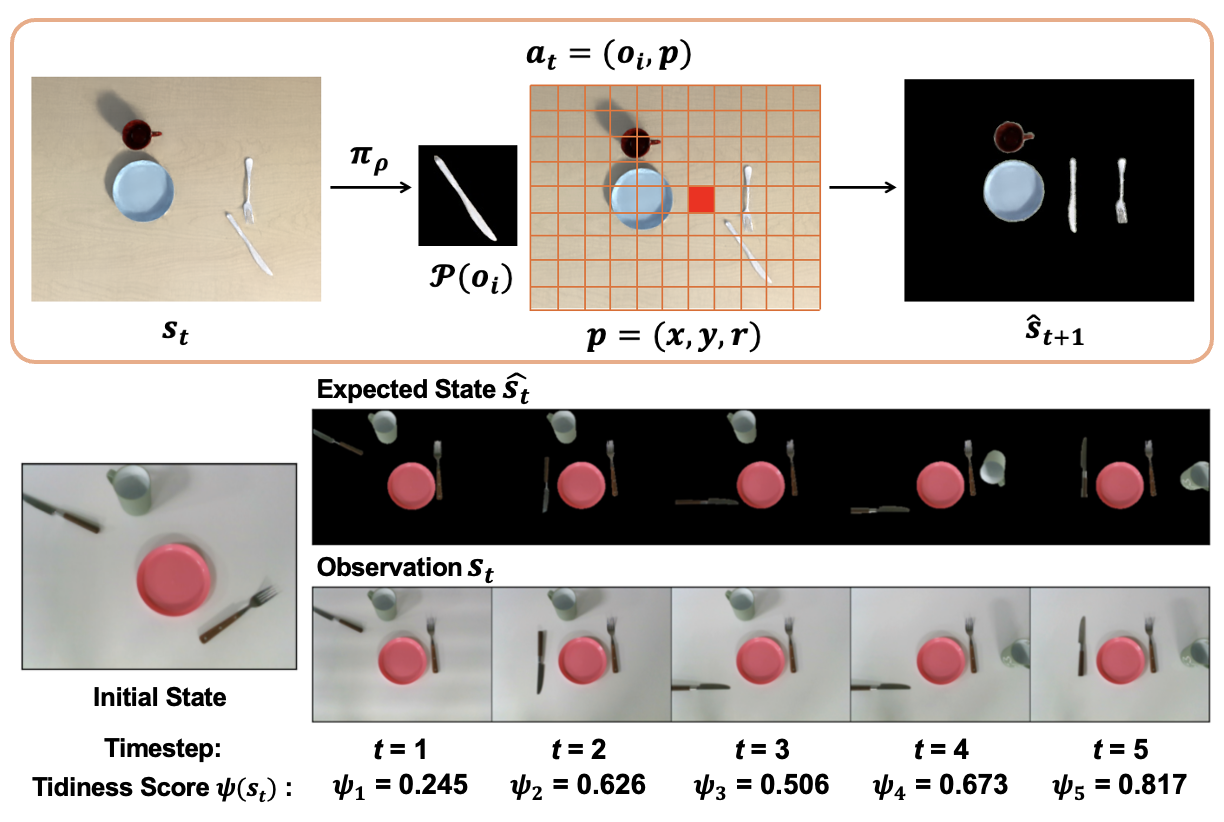}
\end{center}
\vspace*{-5mm}
\caption{
The upper figure illustrates the process of next state prediction by directly moving object patches.
The lower figure depicts a sequence of TSMCTS evaluations in the real world.
The top row presents the predicted states $\hat{s}_{t}$ by moving image patches from the previous states. 
The bottom row displays the observed states $s_t$.
$\psi_t$ denotes the tidiness score of each state $s_t$.
}
\vspace*{-5mm}
\label{fig:dynamics}
\end{figure}

We utilize Monte Carlo Tree Search (MCTS) as a high-level planner.
MCTS is a search algorithm to solve decision-making processes for deterministic problems.
For the implementation of MCTS, it is necessary to recognize the dynamics from a state $s_t$ to the next state $s_{t+1}$ after an action $a_t$ is performed. 
Instead of using a separate simulator to get the predicted next state $\hat{s}_{t+1}$, 
our method generates $\hat{s}_{t+1}$ by directly moving each object's image patch on the initial RGB image.
Although $\hat{s}_{t+1}$ obtained through this method may differ from the state achieved by physically performing a pick-and-place action, 
we demonstrate that this approach achieves an 85\% success rate in real world experiments, indicating its robustness despite potential errors.
Figure \ref{fig:dynamics} shows the process of next state prediction and the sequence of expected and actual states during real world evaluation.

Our high-level policy leverages the trained tidiness discriminator and tidying policy to guide MCTS in finding the most efficient action sequence for tabletop tidying up. 
The overall inference process of TSMCTS is shown in Figure \ref{fig:overview}-b.
For each timestep $t$, the agent receives the state which consists of the current RGB-D images.
Then TSMCTS builds a search tree with the root node representing the current state $s_t$.
Starting from the initial tree, TSMCTS repeats the following four steps $K$ times to complete the tree: Selection, Expansion, Simulation, and Backpropagation.

\textbf{Selection:} 
Starting from the root node, TSMCTS selects child nodes until it reaches a leaf node. 
At each node $s$, TSMCTS selects an action $a$ based on the UCT function, given as follows:  
\begin{small}
\begin{align}
U(s, a) = \frac{Q(s)}{N(s,a)} + c \sqrt{\frac{2\log{N(s)}}{N(s, a)}},
\end{align}
\end{small}where $c$ is the exploration term.  
$N(s)$ denotes the number of visits to node $s$, and $N(s,a)$ denotes the number of times action $a$ has been executed at node $s$.
$Q(s)$ denotes the cumulative reward of node $s$, where the reward is assigned during the Backpropagation step.

\textbf{Expansion:} 
If TSMCTS reaches a leaf node $s_{leaf}$, it adds a child node to the tree.
To expand the tree at the leaf node, we use the trained tidying policy $\pi_\rho$ to sample actions from the action space, $a \sim \pi_\rho(\cdot | s_{leaf} )$.
Here, the action $a$ represents a pick-and-place action, $a=(o, p)$.
As mentioned above, we create the new child node $s_{new}$ by moving the image patch of object $o$ to the position $p$ in the RGB image of $s_{leaf}$.

\textbf{Simulation:} 
We leverage the trained tidiness discriminator to predict the expected value of the expanded node $s_{new}$ as $V(s_{new}) = \Psi_\theta (s_{new})$,
where $\Psi_\theta$ denotes the tidiness discriminator.
Additionally, we obtain the outcome $z(s_{new})$ from a random rollout by executing $\pi_\rho$ until the terminal step $T_{rollout}$.
The outcome $z(s_{new})$ is set to 1 if the final state of the rollout is fully tidied up and 0 otherwise.

\textbf{Backpropagation:} 
TSMCTS backpropagates Q-value updates from the newly expanded node $s_{new}$ back to the root node.  
For each node $s$ and action $a$ along the path, the Q-value updates are performed as follows:
\begin{equation}
\small
\begin{aligned}
&N(s) \leftarrow N(s) + 1, \\
&N(s,a) \leftarrow N(s,a) + 1, \\
&Q(s,a) \leftarrow Q(s,a) + (1-\lambda) V(s_{new}) + \lambda z(s_{new}).
\end{aligned}
\end{equation}
We use $\lambda=0.3$, where $\lambda$ denotes the mixing parameter.

After the tree search is completed, TSMCTS selects the most visited child node of the root node as the best action. 
Then, the high-level action is converted into low-level actions by the low-level planner to control the robot.

\section{EXPERIMENTS}

\subsection{Evaluation of Tidiness Discriminator}
We evaluate whether the trained tidiness discriminator generalizes well to unseen objects and unseen configurations beyond the training data. 
We divide the 224,225 tidying data into 162,000 training data and 62,225 validation data. 
The validation data contains unseen objects and templates from the training data. 
To determine whether a scene is fully tidied up, we define a tidiness threshold $\xi$, ranging from 0 to 1.
During the experiments, a task is considered successful if the tidiness score exceeds $\xi$. 
The tidiness threshold is crucial for determining success - 
lowering $\xi$ increases recall by classifying more scenes as well-tidied but reduces precision due to more false positives. 

\begin{figure}[t!]
\begin{center}
\includegraphics[width=0.48\textwidth]{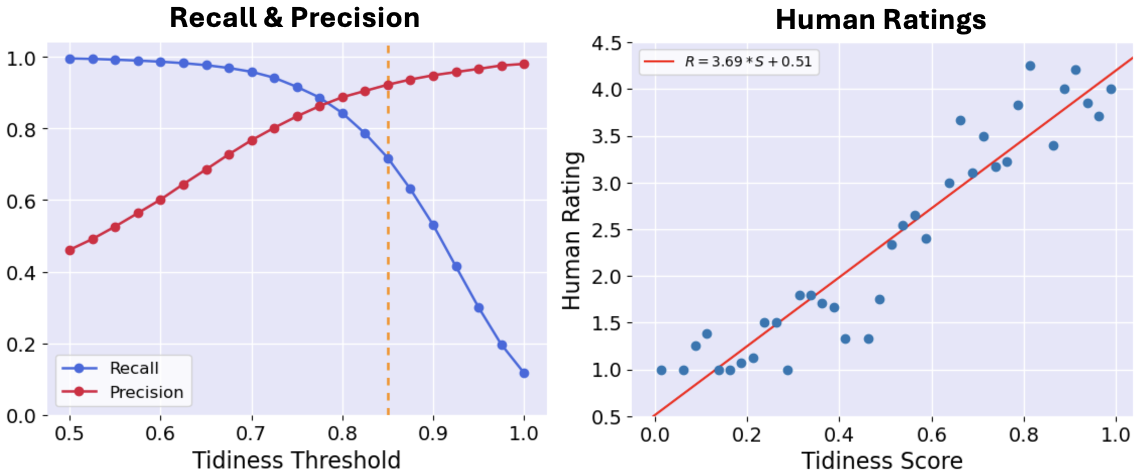}
\end{center}
\vspace*{-4.5mm}
\caption{
The left graph shows the recall and precision measured according to the tidiness threshold.
The orange dashed line represents the tidiness threshold used in the experiments, $\xi=0.85$.
The right graph shows the distribution of human evaluated ratings according to the tidiness scores.
We divide the tidiness score range from 0 to 1 into 40 intervals and average the ratings of scenes within each interval.
}
\vspace*{-6mm}
\label{fig:score}
\end{figure}

To determine an appropriate threshold, we analyze the classification performance of the tidiness discriminator across varying tidiness threshold values.
The recall and precision measured on the validation set as functions of the tidiness threshold are illustrated in Figure \ref{fig:score}.
Additionally, we conduct a human evaluation to ensure that the tidiness threshold aligns with human perceptions of tidiness.
We present 20 randomly selected sequences from 50 tidying sequences organized by TSMCTS to 17 participants, asking them to choose the scenes they judge to be tidied up.
For each sequence, we define the tidiness threshold as the lowest tidiness score among the scenes that participants judge to be tidied up.
The average thresholds for each environment are presented in Table \ref{table:threshold}.
As a result, people judge that the table is tidied up at an average tidiness score of 0.8486. 
Looking at the environment, the threshold for the Dining table is higher at 0.9017 compared to other environments.
This appears to be because people consider the arrangement more organized when the tableware and cutlery are placed according to their functional uses.
Based on these results, we determine that a threshold of 0.85 is an appropriate value.
The trained tidiness discriminator achieve a recall of 71.8\% and a precision of 92.2\% on the validation data using this threshold.

We conduct another human evaluation to verify how well the trained tidiness score reflects the actual perception of tidiness by humans. 
We sample 10 scene data for each tidiness score interval from 0 to 1 at 0.1 intervals, resulting in a total of 100 scenes.
Then, we ask 17 participants to view 15 randomly selected scenes from the 100 scenes and rate the degree of tidiness on a scale of 1 to 5. 
The correlation between the tidiness score and human ratings is shown in Figure 6.
We observe a strong positive correlation between the tidiness score and human ratings.
We also find a tendency for the variance in human ratings to increase as the tidiness score rises. 
This is likely because the standards for tidiness are highly subjective and vary from person to person.

\subsection{Simulation Experiments}
We use the PyBullet simulator for the simulation experiments. 
In the simulator, a workspace table and a UR5 robot are set up. 
As the initial states, random objects are spawned on the table in random positions and orientations.
We use 3D object models from the YCB and HouseCat6D datasets, along with 10 additional object models and four extra object categories not included in the training set of the TTU dataset. 

We evaluate TSMCTS in simulation across five environments by adding, Mixed, a mixed table environment to the original four: Coffee table, Dining table, Office desk, and Bathroom.
For each environment, we tested 150 scenarios with varying object compositions and initial placements.
We use the tidiness threshold 0.85 defined in the previous section for the success criteria.
A failure is noted if objects are placed outside the workspace, collide and overlap, or if tidying is not completed within 10 steps.
We measured the tidying success rate, the tidiness score of the final state, and the number of steps taken.
The experimental results are presented in Table \ref{table:tsmcts_eval_simul}. 
In the Coffee table environment, which has a diverse and complex set of objects, a success rate of 79.3\% and an average tidiness score of 0.889 are achieved, while in the Dining Table environment, which has more standardized object templates, a higher success rate is observed compared to other settings. 
Additionally, high success rates and tidiness scores are also achieved in mixed object configurations, which are not part of the training data. 
TSMCTS demonstrates its ability to successfully find arrangements that meet tidiness conditions across a variety of environments and object configurations.

\begin{figure*}[htb!]
\begin{center}
\includegraphics[width=0.93\textwidth]{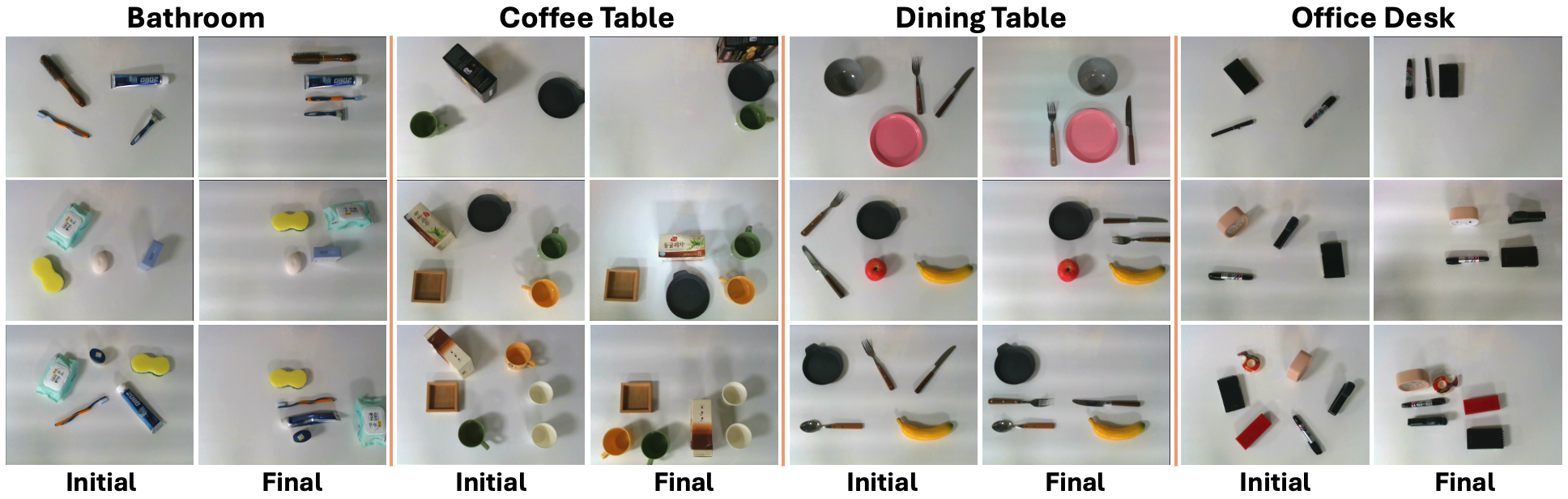}
\end{center}
\vspace*{-5mm}
\caption{
Examples of tidying up across various object sets. 
Starting from the initial configurations, TSMCTS successfully tidied up the tables.
The demonstration includes various objects such as apples, bananas, clocks, razors, and scotch tape, which are not included in the TTU dataset.
}
\vspace*{-5mm}
\label{fig:tsmcts_result}
\end{figure*}

\begin{table}[t!]
\centering
\caption{
Tidiness Threshold measured by Human Evaluation
}
\label{tab:comprehensive_performance_evaluation}
\setlength{\tabcolsep}{15pt} 
\renewcommand{\arraystretch}{1}
\vspace*{-2mm}
\begin{tabular}{c|c}
\toprule
Environment & Tidiness Threshold \\
\midrule
Coffee Table & 0.8223 $\pm$ 0.1230\\
Dining Table & 0.9017 $\pm$ 0.0895\\
Office Desk & 0.8310 $\pm$ 0.0892\\
Bathroom & 0.8632 $\pm$ 0.0799\\
\midrule
Average  & 0.8486 $\pm$ 0.0423\\
\bottomrule
\end{tabular}
\vspace*{-2mm}
\label{table:threshold}
\end{table}

\begin{table}[t!]
\centering
\caption{
TSMCTS Evaluation in the Simulation
}
\vspace*{-2mm}
\label{tab:comprehensive_performance_evaluation}
\setlength{\tabcolsep}{6pt} 
\renewcommand{\arraystretch}{1}
\begin{tabular}{c|ccc}
\toprule
Environment & Success Rate $\uparrow$ & Tidiness Score $\uparrow$ & Length $\downarrow$  \\
\midrule
Coffee Table     & 79.3\% & 0.889 & 4.631 \\
Dining Table    & 92.7\% & 0.904 & 5.144  \\
Office Desk    & 90.7\% & 0.905 & 4.778  \\
Bathroom & 89.3\% & 0.902 & 4.626  \\
Mixed    & 90.7\% & 0.907 & 4.620  \\
\midrule
Average    & 88.5\% & 0.901 & 4.760  \\
\bottomrule
\end{tabular}
\vspace*{-6.5mm}
\label{table:tsmcts_eval_simul}
\end{table}

For baseline comparisons, we evaluate TSMCTS comparing StructFormer \cite{StructFormer} and StructDiffusion \cite{StructDiffusion}.
Both StructFormer and StructDiffusion are algorithms that find arrangements matching given conditions, based on language tokens related to goals. 
While their setup is different from ours, both studies include tasks for organizing a dining table, so we conduct comparative experiments exclusively in the Dining table setting. 
Additionally, we perform an ablation study on the tidiness discriminator by comparing TSMCTS with TSMCTS-binary. 
TSMCTS-binary utilizes a tidiness discriminator trained with binary labels from the TTU dataset, where completely tidied scenes are labeled as 1, and all other scenes are labeled as 0.

We conduct a human evaluation to determine how closely the results tidied by each algorithm approximate human-perceived tidiness.
We ask 17 participants to view 30 randomly selected scenes and move objects as much as they desire until they achieve a tidiness that satisfy them. 
For this purpose, we collect 20 final tidied scenes from each algorithm and include scenes during the tidying process, preparing a total of 160 scene data. 
We obtain a segmentation mask for each scene, which allow participants to move object patches using keyboard controls. 
Movements are set to adjust 1 cm per step in any direction, and rotations could be adjusted by 10 degrees clockwise or counterclockwise. 
We measure the cumulative object movement distance, cumulative object rotation angle, and cumulative number of keyboard operations for each scene. 
Additionally, the participants are asked to fill out a NASA task load index (NASA-TLX) \cite{nasatlx} form for each scene.
NASA-TLX form is used to evaluate the workload of a task, which divides the total workload into six subjective subscales. 
In this paper, we focus on three subscales that are most relevant to our task: mental demand, own performance, and frustration level.
The results of human evaluation are presented in Table \ref{table:human_eval_simul}. 
TSMCTS shows the lowest total object movement distance at 57.2cm and fewest keyboard operations at 102.9. 
However, the NASA-TLX score is lowest for TSMCTS-Binary at 26.47, while StructDiffusion and StructFormer achieve lower total rotation angles than TSMCTS.
This suggests that point cloud-based algorithms perform better in finding the appropriate orientation for each object.
Low NASA-TLX scores for TSMCTS and TSMCTS-Binary indicate that participants feel less task load when tidying the table, implying less effort or stress is required to achieve a tidied arrangement that meets human standards. 
The minimal movement distance and keyboard operations for TSMCTS suggest it positions objects closest to what people consider a tidy arrangement.

\begin{table}[t!]
\centering
\caption{
Human Evaluation in Simulation Experiments 
}
\vspace*{-1.5mm}
\label{tab:comprehensive_performance_evaluation}
\setlength{\tabcolsep}{2pt} 
\renewcommand{\arraystretch}{1}
\hspace*{2mm}
\begin{tabular}{c|cccc}
\toprule
\multirow{2}{*}{Methods} & \multirow{2}{*}{Distance $\downarrow$} & \multirow{2}{*}{Rotation $\downarrow$} & Number of & \multirow{2}{*}{NASA-TLX $\downarrow$} \\
 & &  & Operations &  \\
\midrule
StructFormer \cite{StructFormer}    & 88.2cm & 150.3\textdegree & 143.1 & 37.84  \\
StructDiffusion \cite{StructDiffusion} & 65.3cm & \textbf{101.9\textdegree} & 116.5 & 41.86  \\
TSMCTS-Binary    & 60.3cm & 201.0\textdegree & 104.0 & \textbf{26.47}  \\
TSMCTS           & \textbf{57.2cm} & 158.1\textdegree & \textbf{102.9} & 27.06  \\
\bottomrule
\end{tabular}
\vspace*{-6mm}
\label{table:human_eval_simul}
\end{table}

\subsection{Real Robot Experiments}
We use a Universal Robots UR5 mounted with a Robotiq 2F-85 Gripper at the end effector for real robot experiments. 
An Intel RealSense D435 camera is mounted on the wrist of UR5 to capture RGB-D images in $480\times640$ resolution.
At each timestep, our agent receives RGB and depth images from the mounted camera at the fixed view point. 

We evaluate TSMCTS across four environments: Coffee table, Dining table, Office desk, and Bathroom.
We conduct tidying scenarios with five different object configurations in each environment and measure the tidiness score of the final scene for each scenario. 
If the tidiness score exceeds the tidiness threshold within 10 timesteps, the scenario is considered a success. 
However, if tidying is not achieved within 10 timesteps, if objects become entangled and a grasping point cannot be found, or if an object leaves the workspace, the scenario is considered a failure.
The results of the experiments are presented in Table \ref{table:tsmcts_eval_real}, and demonstrations of the final tidied up tables are shown in Figure \ref{fig:tsmcts_result}.
TSMCTS achieves an average tidiness score of 0.897 and a success rate of 85\% across a total of 20 scenarios.
In real experimental scenarios, objects not shown in the TTU dataset are also included in the configurations.
These results demonstrate that TSMCTS can robustly tidy up even in scenarios with diverse and complex object compositions.

\begin{table}[t!]
\centering
\caption{
TSMCTS Evaluation in the Real World
}
\vspace*{-1.5mm}
\label{tab:comprehensive_performance_evaluation}
\setlength{\tabcolsep}{4pt} 
\renewcommand{\arraystretch}{1}
\begin{tabular}{c|ccc}
\toprule
Environment & Success Rate $\uparrow$ & Tidiness Score $\uparrow$ & Length $\downarrow$  \\
\midrule
Coffee Table     & 100\% & 0.894 & 4.4 \\
Dining Table    & 80\% & 0.840 & 3.8  \\
Office Desk    & 80\% & 0.945 & 7.0  \\
Bathroom & 80\% & 0.909 & 5.6  \\
\midrule
Average    & 85\% & 0.897 & 5.1  \\
\bottomrule
\end{tabular}
\vspace*{-6mm}
\label{table:tsmcts_eval_real}
\end{table}

\begin{figure}[tb!]
\begin{center}
\includegraphics[width=0.43\textwidth]{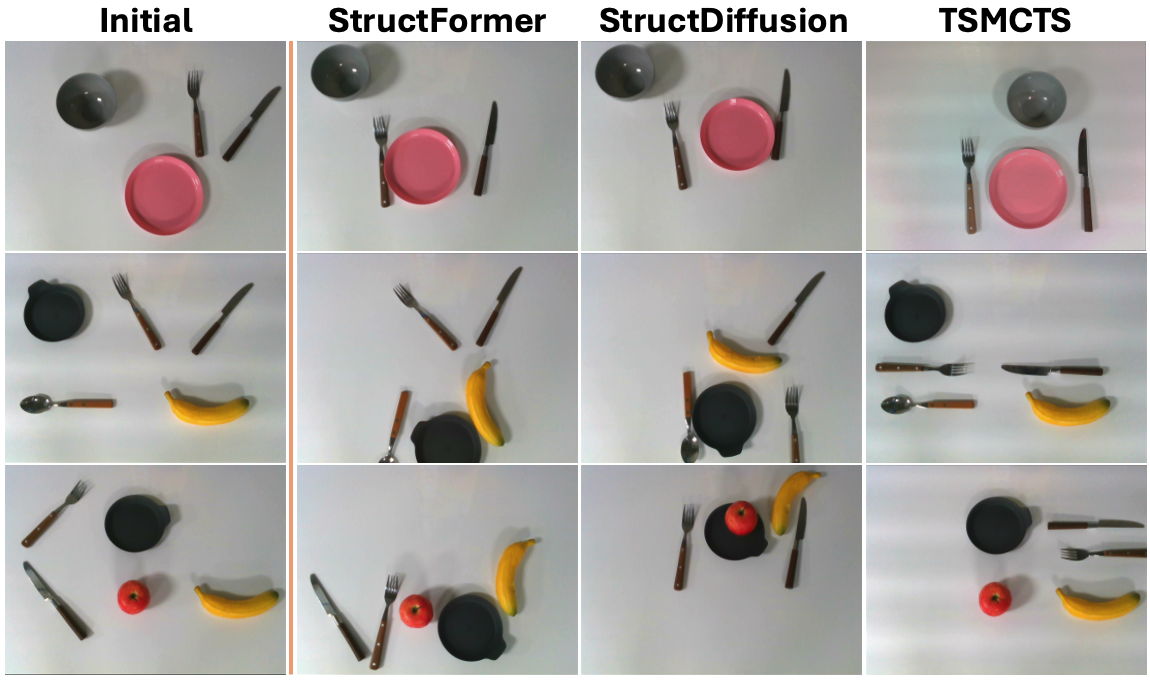}
\end{center}
\vspace*{-5mm}
\caption{
Examples of tidying up using various methods in the real world. We evaluated StructFormer, StructDiffusion, and TSMCTS on a Dining table setup, starting from same initial configurations.
}
\vspace*{-2mm}
\label{fig:result_compare}
\end{figure}

For baseline comparisons, we evaluate TSMCTS comparing StructFormer and StructDiffusion.
We conduct comparative experiments solely in the Dining table setting, similar to the simulation experiment.
We measure average scenario length and average number of collisions for the five scenarios. 
StructFormer and StructDiffusion are considered successful if they arrange objects according to the intended positions without any collisions.
For TSMCTS, the success condition is the same as the previous experiment.
The results of the experiments are presented in Table \ref{table:realworld_compare}, 
and the final tidied up tables are shown in Figure \ref{fig:result_compare}.
In a simple dining table setup with one plate, fork, and knife, StructFormer and StructDiffusion demonstrate good tidying performance. 
However, in complex configurations with multiple forks or knives, StructFormer and StructDiffusion often fail to find appropriate positions for all objects, leading to overlaps and collisions. 
TSMCTS, on the other hand, is able to find tidied arrangements even with complex configurations, 
demonstrating its diverse and robust tidying capabilities.

\begin{table}[t!]
\centering
\caption{
Real World Evaluation on the Dining Table Environment
}
\vspace*{-1.5mm}
\label{tab:comprehensive_performance_evaluation}
\setlength{\tabcolsep}{6pt}
\renewcommand{\arraystretch}{1}
\begin{tabular}{c|ccc}
\toprule
Methods & Success Rate $\uparrow$ & Length $\downarrow$ & Collisions $\downarrow$  \\
\midrule
StructFormer \cite{StructFormer} & 40\%  & 3.6 & 0.6 \\
StructDiffusion \cite{StructDiffusion} & 60\%  & \textbf{3.0} & 0.4 \\
TSMCTS & \textbf{80\%} & 3.4 & \textbf{0.2} \\
\bottomrule
\end{tabular}
\vspace*{-5mm}
\label{table:realworld_compare}
\end{table} 

\section{CONCLUSION}
In this paper, we have introduced the TSMCTS framework, a tidiness score-guided Monte Carlo tree search for tabletop tidying up.
TSMCTS is a framework that uses a tidiness discriminator to assess current and future table tidying states, generates a search tree according to the tidying policy, and finds the optimal arrangement for tidying up. 
To train the tidiness discriminator and tidying policy, we have collected the TTU dataset, a structured dataset that includes tidying sequence data across various environments.
We have shown experimental results that TSMCTS has robust tidying capabilities across various object configurations including unseen objects and duplicate objects.
In addition, we have successfully transferred TSMCTS to the real world without any transferring efforts. 
Despite the satisfactory results, there also exist limitations in the proposed method. 
Since TSMCTS assumes a 2D arrangement, it cannot perform tidying that involves stacking objects in layers.
Additionally, 
the tidiness discriminator relies on visual information,
which often leads to a lack of consideration for the functional uses of objects.
In future work, we plan to leverage large language models (LLMs) as guidance to better handle ambiguous cases and resolve scenarios where the functional use or arrangement of objects is unclear.





\bibliographystyle{IEEEtran} 
\vspace*{-1mm}
\bibliography{ref} 

\end{document}